\documentclass[runningheads]{llncs}
\usepackage{booktabs}
\usepackage{algorithm} 
\usepackage[noend]{algpseudocode}
\usepackage{multirow}
\usepackage{xcolor,colortbl}
\usepackage{graphicx}
\usepackage{framed}
\usepackage{tikz}
\usepackage{calc}
\usepackage{amsmath}
\usepackage{amssymb}
\usepackage{hyperref}
\usepackage{subcaption}
\usepackage{comment}
\usepackage{cleveref}
\usepackage{ragged2e}
\graphicspath{{./pictures/}}
\makeatletter

\newcommand{\@todobase}[3]{\colorbox{#1}{\textbf{#2}}\hspace{5pt}\textcolor{#1}{#3}}
\newcommand{\@todo}[2][red]{\@todobase{#1}{ToDo:}{#2}}
\newcommand{\@todostar}[3][red]{\@todobase{#1}{#2}{#3}}
\newcommand{\todo}{\@ifstar{\@todostar}{\@todo}}

\newcommand{\head}[1]{\textnormal{\textbf{#1}}}

\newcounter{@i}

\newcommand{\nxt}{\stepcounter{@i}\the@i.}

\newcommand{\secref}[1]{\hyperref[#1]{Section~\getrefnumber{#1}}}
\newcommand{\tabref}[1]{\hyperref[#1]{Tab.~\getrefnumber{#1}}}
\newcommand{\figref}[2][\empty]{%
\if&#1&%
\hyperref[#2]{Fig.~\getrefnumber{#2}}%
\else%
\hyperref[#2]{Fig.~\getrefnumber{#2}#1}%
\fi%
}

\makeatother

\begin{document}
\title{Early prediction of the risk of ICU mortality with Deep Federated Learning}
%
%
\author{Korbinian Randl\inst{1}\orcidID{0000-0002-7938-2747} \and
Núria Lladós Armengol\inst{1}\orcidID{0000-0002-8584-5058} \and
Lena Mondrejevski\inst{1}\orcidID{0000-0002-1790-3842} \and
Ioanna Miliou\inst{1}\orcidID{0000-0002-1357-1967}}
\authorrunning{K. Randl, N. Lladós Armengol et al.}
%
\institute{Dept. of Computer \& Systems Sciences,
    Stockholm University, Stockholm, Sweden \\
    \email{k.randl@web.de, nuriallados@hotmail.com, \{lena.mondrejevski,ioanna.miliou\}@dsv.su.se}
}
\maketitle
\setcounter{footnote}{0}%
\vspace{-6mm}
\begin{abstract}
Intensive Care Units usually carry patients with a serious risk of mortality. Recent research has shown the ability of Machine Learning to indicate the patients' mortality risk and point physicians toward individuals with a heightened need for care. Nevertheless, healthcare data is often subject to privacy regulations and can therefore not be easily shared in order to build Centralized Machine Learning models that use the combined data of multiple hospitals. Federated Learning is a Machine Learning framework designed for data privacy that can be used to circumvent this problem. In this study, we evaluate the ability of deep Federated Learning to predict the risk of Intensive Care Unit mortality at an early stage.
We compare the predictive performance of Federated, Centralized, and Local Machine Learning in terms of AUPRC, F1-score, and AUROC. Our results show that Federated Learning performs equally well as the centralized approach and is substantially better than the local approach, thus providing a viable solution for early  Intensive Care Unit mortality prediction. In addition, we show that the prediction performance is higher when the patient history window is closer to discharge or death. Finally, we show that using the F1-score as an early stopping metric can stabilize and increase the performance of our approach for the task at hand.


\end{abstract}
\vspace{-8mm}
\section{Introduction}
\vspace{-2mm}
Intensive Care Units~(ICUs) usually treat patients with a heightened mortality risk. A study conducted on patients admitted to 167 ICUs from 17 European countries during four weeks in 2011, and 2012 registered that, out of $5,834$ patients, $1,113\,(19\,\%)$ died in the ICU alone and a total of $1,397$ patients $(24\%)$ died in the whole hospital~\cite{Capuzzo2014}. Mortality increases even further when considering patients admitted to an ICU during 2020 and 2021. Among $1,686$ patients admitted to the ICU with COVID-19, the mortality rate was $30\%$~\cite{covid2022}.
\newline\indent The ICU is also one of the units where medical errors are most likely to occur, given the complexity of care. ICU patients are severely ill and usually subject to multiple complex interventions and treatments, thus leading to a high risk of an adverse outcome. 
In order to enable clinicians to take action and prevent such an outcome, the risk of patients' mortality has to be predicted not only as accurately as possible but also as early as possible; we refer to this concept as \textit{early ICU mortality prediction}.
ICUs are therefore equipped with advanced diagnostic and therapeutic resources in order to enable quick response to changes in patients' health. Traditionally, hospitals use the collected Electronic Health Records~(EHRs) to assess individual mortality risks in the ICU with the help of scores like APACHE~\cite{Knaus1991} and SAPS~\cite{Gall1993}.
%
\newline\indent Recent research in the field of Machine Learning (ML) has shown improvements in these scores in terms of predictive performance \cite{Awad2020,Johnson2017}. The results of Awad et al.~\cite{Awad2020} clearly showcase the performance improvement of classical ML methods over clinical severity scores for early ICU mortality prediction. Similarly, Johnson and Mark \cite{Johnson2017} show that classical ML models trained using data from the first $24\,\mathrm{h}$ of the ICU stay can outperform clinical scores.
\newline\indent These ML models rely on big amounts of data in order to learn correlations between current patient data and their risk of mortality within a specific time window in the future. 
Conventionally, ML approaches use all the available data to train the model in a centralized manner. 
However, as patient data are usually subject to privacy regulations, like the European GDPR, the data cannot simply be shared between hospitals to train Centralized ML~(CML) models. Alternatively, locally available data could be used by each hospital to set up its own independent Local ML~(LML) early warning system. However, this paradigm could suffer from a lack of sufficient training data and would not consider the heterogeneity of patients across multiple medical centers.
\newline\indent A promising alternative to LML or CML is presented by Federated Learning~(FL). Instead of exchanging data, this ML framework relies on training many local models of identical structure at the location of the data, which are then combined into a global model. In our case, the data are stored at different hospitals, which we refer to as \textit{clients}. Apart from ensuring privacy by design, FL enables parallel training using the clients as computational units \cite{McMahan2016,Shokri2015}.
\newline\indent FL can be perfectly integrated into hospital facilities that store patients' information in the form of EHRs. An advantage of EHRs is that data are collected in the form of Multivariate Time-Series~(MTS), which means that each feature consists of a stream of values changing over time. Deep Learning~(DL) architectures like Recurrent Neural Networks~(RNNs) represent powerful tools for dealing with this kind of data, 
as they also take into account the history encoded by the time series data. One major challenge when training a DL model is to find a trade-off between training the neural network enough to learn the features of the training set and  stopping before it overfits the training data. One possible technique to find the optimal number of training epochs is Early Stopping~(ES). 
In case of overfitting, the validation performance of the model 
begins to degrade, and the training process stops. Different metrics, such as loss or F1-score, can be used to trigger ES. Their suitability depends on the task.
\newline\indent Some approaches in literature use RNNs for binary classification on MTS from the ICU; however, they do not tackle the problem in combination with FL. 
Pattalung et al.~\cite{Pattalung2021} compare different RNN architectures to create benchmark values for \textit{ICU mortality prediction} tasks on different publicly accessible databases. The authors take MTS data over a $48\,\mathrm{h}$ period in order to predict whether a patient dies at the end of this period.
Ge et al.~\cite{Wendong2018} combine Logistic Regression and LSTM for \textit{early ICU mortality prediction} after $48\,\mathrm{h}$ of ICU admission. Their results show that DL achieves higher accuracy than Logistic Regression in identifying patients at high risk of death.
\newline\indent Despite the benefits of using an FL setup, few papers address the problem of mortality prediction with FL.
In this study, we built upon the research of Mondrejevski et al.~\cite{Mondrejevski2022} that propose FLICU, an FL workflow for mortality prediction tasks in the ICU. However, their paper does not focus on predicting ICU mortality at an early stage but it is rather a retrospective study.
\newline\indent This paper tackles the current limitations by combining FL with \textit{early prediction} of ICU mortality. The main contributions are summarized below: 
\vspace{-1mm}
\begin{enumerate}
\item We propose a workflow for predicting early ICU mortality using deep FL.
\item We analyze the predictive performance of our proposed solution on different time~windows and data cohorts.
\item We compare the predictive performance of the FL setup (with $2$, $4$, and $8\,$clients) against CML and LML.
\item We compare two early-stopping criteria during training: loss and F1-score.
\end{enumerate}
\vspace{-4mm}




\section{Method}
\vspace{-2mm}
In this paper, we propose a workflow for early ICU mortality prediction that comprises three main phases: \textbf{(i)}~\nameref{sec:data-preparation}, \textbf{(ii)}~\nameref{sec:window-selection}, and \textbf{(iii)}~\nameref{sec:deep_learning}. A schematic view of our workflow is shown in \figref[a]{fig:method}.
\vspace{-5mm}

\begin{figure}[h]
\begin{minipage}{0.4\textwidth}%
\begin{tikzpicture}%
\draw (0,0) node[anchor=north east]{\includegraphics[trim = 0.9cm 2.6cm 0.9cm 0.4cm, clip, width=\textwidth]{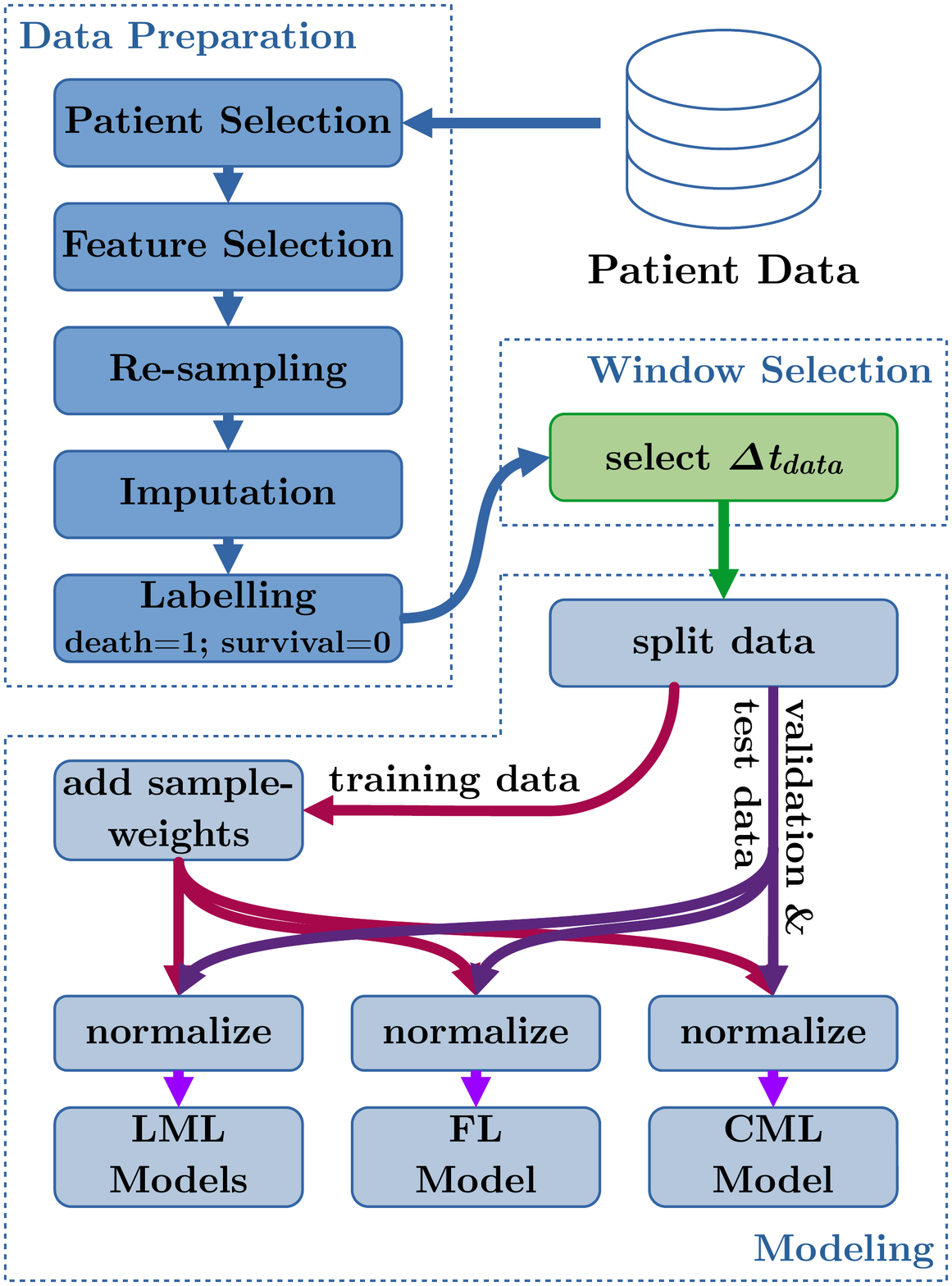}};%
\draw (-.2,0) node[anchor=north east]{\footnotesize{\color{gray}(a)}};%
\end{tikzpicture}%
\end{minipage}%
\hspace{5pt}%
\vrule%
\hspace{5pt}%
\begin{minipage}{0.56\textwidth}%
\noindent\begin{tikzpicture}%
\draw (0,0) node[anchor=north west]{\includegraphics[trim = 0.1cm 0.0cm 0.1cm 0.2cm, clip, width=.98\textwidth]{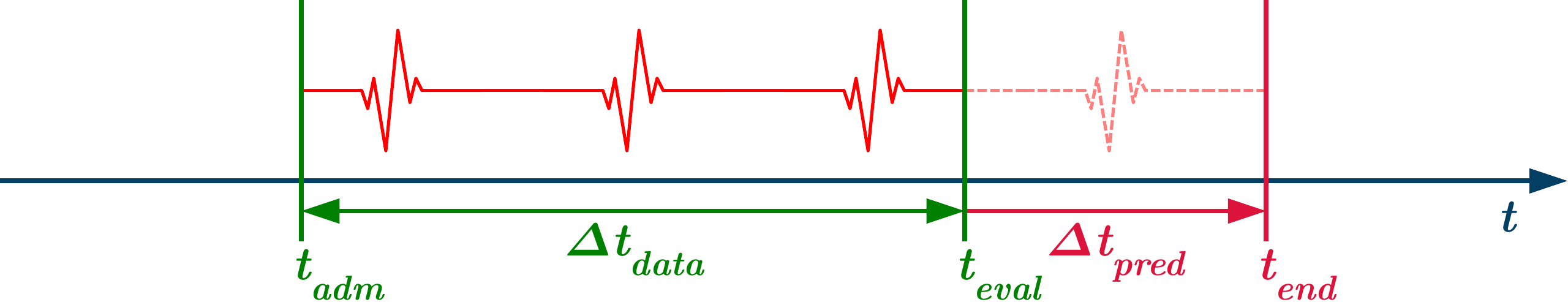}};%
\draw (0,0) node[anchor=north west]{\footnotesize{\color{gray}(b)}};%
\end{tikzpicture}%

\noindent\justifying\tiny{Patient data is available from the admission to the ICU $t_{adm}$ to death or discharge from the ICU $t_{end}$. Training data is sampled during $\Delta t_{data}$ (from $t_{adm}$ to $t_{eval}$). This leaves a prediction window $\Delta t_{pred}$, during which data is unknown to the model.}%

\vspace{5pt}%
\hrule%
\vspace{5pt}%

\noindent\begin{tikzpicture}%
\draw (0,0) node[anchor=north west]{\includegraphics[trim = 0.1cm 0.1cm 0.1cm 0.5cm, clip, width=.98\textwidth]{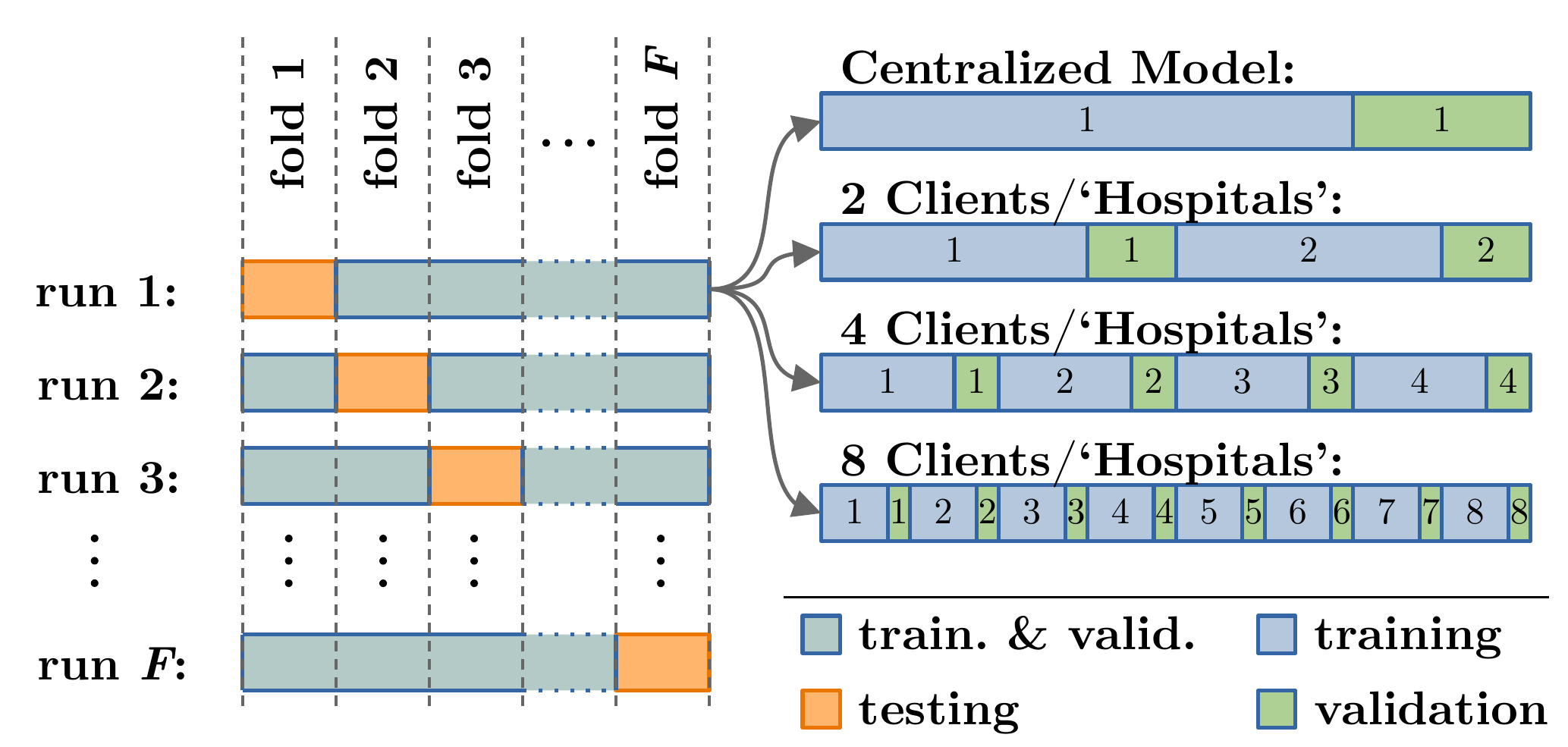}};%
\draw (0,0) node[anchor=north west]{\footnotesize{\color{gray}(c)}};%
\end{tikzpicture}%

\noindent\justifying\tiny{Data are split into $F$ equally sized folds. Over $F$ iterations, each of the folds is used $F-1$ times for training and validation and once for testing. For FL and LML, we further split the combined training and validation data into the number of clients. These partitions are, in turn, divided into training and validation data.}%
\end{minipage}%
\caption{(a)$\,$Schematic Workflow, (b)$\,$Time-window representation (c),$\,$Data splits.}%
\label{fig:method}%
\end{figure}%

\subsection{Problem Formulation}\label{sec:problem_formulation}
\vspace{-1mm}%
Our basic research problem can be formulated as a binary classification problem for early ICU mortality prediction: given a patient cohort~$D$ that consists of $n$~patients, we aim to estimate the real class label $y_i\in[0,1]$ for each patient~$i\in[0,1,\dotsc,n]$ by predicting a label $\hat{y}_i$. The label $y_i$ denotes whether patient~$i$ died or survived the ICU stay and $\hat{y}_i$ indicates the patient's predicted mortality risk. The label~$\hat{y}_i$ is estimated based on an MTS feature stream $X_i(t)=[x_{i0}(t),x_{i1}(t),\dotsc,x_{im}(t)]$, that consists of $m$~features (e.g. vital signs or laboratory values). Each $x_{ij}(t),~j\in[0,1,\dotsc,m]$ represents the value of a univariate time series at time~$t$. Since the main focus of this study is \textit{early prediction}, the observations $X_i(t)$ that are used are collected only during the first hours of the patient's ICU stay. 

Furthermore, we assume that the patients are randomly distributed over $K$ ICUs/hospitals. $D_k$ is the local cohort of patients (i.e., the set of locally available $X_i(t)$) in each hospital~$k$. Thus, $D=D_0~\cup~D_1~\cup~\dotsc~\cup~D_K$ is the global dataset of $n$~patients with $D_a\cap D_b = \emptyset, \forall (a, b) \in [1, K]\times[1, K]$. 
In this paper, we compare the efficiency of CML, LML, and FL on this problem.
\vspace{-3mm}

\subsection{Data Preparation}\label{sec:data-preparation}
\vspace{-1mm}%
In the first phase of our workflow, we are preparing the data for the predictive modeling, following five steps (based on the FLICU-workflow \cite{Mondrejevski2022}): 

\textbf{(i)}~\textbf{Patient selection}: Initially, we select patients with an ICU stay and filter the cohort according to the criteria below:
\vspace{-1.5mm}
\begin{enumerate}
    \item We dismiss all but the first ICU stay of each patient.
    \item We dismiss patients with data recorded for less than $\Delta t_{min}$. 
    \item We dismiss patients staying longer than $\Delta t_{max}$.
\end{enumerate}
\vspace{-1.5mm}
The windows $\Delta t_{min}$ and $\Delta t_{max}$ allow us to define the patient cohorts. $\Delta t_{min}$ denotes the minimum length of MTS data per patient. Additionally, it guarantees that all patients will have at least the required history length. The upper bound $\Delta t_{max}$ limits the prediction window in order to dismiss patients that stay in the ICU for an extensive period of time.

\textbf{(ii)}~\textbf{Feature selection}: Consecutively, we extract vital signs and lab values in the form of MTS. We extract vitals explicitly connected to ICU stays and consider the time of the first vital of each patient to be the time of admission to the ICU. If no vitals are recorded for an ICU stay, we use the admission time logged by the hospital staff. Lab values are sampled during the patient's whole ICU stay, from ICU admission to ICU discharge or death.

\textbf{(iii)}~\textbf{Re-sampling}: Vital signs and lab values are re-sampled to fixed length intervals. The length of these intervals depends on the data collection frequency in the hospitals and may differ for vitals and labs. If more than one measurement falls in the same interval, the median is used for data aggregation.

\textbf{(iv)}~\textbf{Imputation}: Missing values are treated with forward, and then backward imputation, starting from the beginning of the ICU stay. Non-observed features for a patient's ICU stay are replaced with $-1$.

\textbf{(v)}~\textbf{Labeling}: Finally, if a patient dies in the ICU, we assign the class label $y_i=1$ (\textit{death}). Otherwise, we assume that the patient survived the ICU and assign the label $y_i=0$ (\textit{discharge}).

\vspace{-2mm}
\subsection{Window Selection}\label{sec:window-selection}
\vspace{-1mm}%
In the second phase of our workflow, we are interested in selecting the patient history window that will be used as input to the predictive models, defined as $\Delta t_{data}\leq \Delta t_{min}$. $\Delta t_{data}$ starts at the time of ICU admission $t_{adm}$ and ends at the time of evaluation $t_{eval}$ (see \figref[b]{fig:method}). Since the main focus of this study is \textit{early ICU mortality prediction}, we aim to predict the label $\hat{y}_i$ ahead of the patient's time of ICU discharge or death, thus $\Delta t_{data}$ is considered to end before the end of the ICU stay $t_{end}$. More precisely, there is a prediction window $\Delta t_{pred} \ge 0$ that starts at $t_{eval}$ and ends at $t_{end}$.
\vspace{-2mm}

\subsection{Modeling}\label{sec:deep_learning}
\vspace{-1mm}%
To avoid bias induced by data partitioning, we use \textit{$F$-fold cross-validation} as the model evaluation technique. We split the data into $F$ partitions, where each split is used as a testing dataset once, while the remaining folds are split into training and validation sets for each client, respectively (see \figref[c]{fig:method}).
To allocate the data points throughout the clients (in LML and FL), we assume $K$ horizontal, stratified splits. This means that the data in each hospital~$k\in[0,1,\dotsc,K]$ has the same number of patients $|D_k|=\frac{n}{K}$ with the same class distribution and that each patient's records are kept in only one hospital at the same time.
Before being passed into the model, all data streams are normalized according to the global minima and maxima found in the available training and validation data.

To deal with class imbalance during training, class weights are added to the training data. This method down-weighs the impact of classes with more examples (in our case: the discharged patients) and increases the importance of classes with fewer data examples during error back-propagation. Thus, applying class weights enables us to achieve optimization-level class balance. The weights $w_c$ for each class $c$ are calculated according to the following formula:
$
w_c = \frac{\text{\# of samples in dataset}}{\text{\# of samples in class }c}
$.
\vspace{1mm}

This setup is aggregated to the three configurations of our predictive model, FL, CML, and LML. As both main gated RNN architectures, LSTM \cite{Hochreiter1997}, and GRU \cite{Cho2014} have shown to perform equally well on tasks comparable to ours \cite{Mondrejevski2022,Pattalung2021}, we focus on the less resource-intensive GRU in this study. 
Our basic DL model architecture (similar to \cite{Mondrejevski2022} and \cite{Pattalung2021}) consists of two parallel input layers, one for the vital signs and one for the laboratory variables, each followed by three recurrent layers of 16~GRUs. Consecutively, we perform batch normalization and combine the resulting outputs using two fully connected layers. 
Finally, we add a sigmoid-layer for estimating the patient risk of ICU mortality with a value between~$0$ and~$1$. Matching to this binary output of the system, we use \textit{binary cross-entropy} as the loss function~$L(\cdot)$.

In CML, the model~$f^{CML}(\cdot,\theta)$ is trained on the whole training data $D$, where $\theta$ is a set of DL weights defining the function of the model. The goal is to find the optimal set of weights~$\theta^{CML}$ that minimizes the error between $y_i$ and $\hat{y}_i=f^{CML}(X_i(t),\theta)$ for a given $X_i(t)$. More formally:
\vspace{-2mm}
\begin{equation}
\theta^{CML}~=~\underset{\theta^*\in\mathbb{R}}{\mathrm{argmin}}~L(D,\theta^*)
\label{equ:global_objective}
\vspace{-2mm}
\end{equation}
After each epoch of training, we evaluate the predictive performance of $f^{CML}(\cdot,\theta)$ on the validation data using a previously selected metric $M(\bf{\hat{y}},\bf{y})$, comparing a vector of predictions~$\bf{\hat{y}}$ with the vector of true class labels~$\bf{y}$. In order to avoid overfitting the training data, we stop the training and reset $\theta$ to the time of the best score~$s^*$ if $s=M(\cdot)$ does not improve for a predefined number of epochs~$P$. We refer to this as early stopping (ES) with patience~$P$.

In LML, each hospital $k$ trains a \textit{local model}~$f^{LML}_k(\cdot,\theta)$, using only the data in the local dataset $D_k$, where $\hat{y}_i=f^{LML}_k(X_i(t),\theta_k)$. As before, the goal is to minimize the prediction error by producing an optimal set of weights~$\theta^{LML}_k$ for each client~$k$:
\vspace{-2mm}
\begin{equation}
\theta^{LML}_k~=~\underset{\theta^*\in\mathbb{R}}{\mathrm{argmin}}~L(D_k,\theta^*)
\label{equ:local_objective}
\end{equation}
Similarly to CML, we calculate local validation scores $s_k=M(\bf{\hat{y}}_k,\bf{y}_k)$ on each client~$k$ in every epoch and monitor them for ES.

For the FL approach, we use a slightly modified version of Federated Averaging (FedAvg) \cite{McMahan2016}:
\textbf{(i)}~First, all \textit{local models} of the $K$ hospitals are initialized with the same set of starting weights $\theta^0$.
\textbf{(ii)}~By performing \textit{local training} for $E$~local epochs, each of the participating hospitals derives an updated set of weight parameters~$\theta_k$. The fraction of participating hospitals per round is $C$. The local objective is similar to Equation~\ref{equ:local_objective}.
\textbf{(iii)}~After local training, a \textit{global model}~$f^{FL}(\cdot,\theta)$ is created by averaging all $\theta_k$ to one set of parameters $\theta$.
\textbf{(iv)}~These averaged weights are then sent back to the hospitals, which overwrite their local weights with the new ones.
\textbf{(v)}~Afterwards, an ES score $s$ is calculated by evaluating the local models on the respective validation set of each client $k$ and then averaging the results: $s=\sum^{K}_{k=1} \frac{n_k}{n} M(\bf{\hat{y}}_k,\bf{y}_k)$. The above is repeated from step~\textbf{(ii)} until the validation scores suggest an optimal set of weights~$\theta^{FL}$ and ES is activated. Here, the objective is to optimize the \textit{global model} $f^{FL}(\cdot,\theta)$ by calculating: 
\vspace{-2mm}
\begin{equation}
\theta^{FL}~=~\underset{\theta^*\in\mathbb{R}}{\mathrm{argmin}}~\sum\limits^{K}_{k=1} \frac{n_k}{n} F_k(\theta^*)
\vspace{-2mm}
\end{equation} 
where
$F_k(\theta^*)=\frac{1}{n_k} L(D_k,\theta^*)$,  $n_k=|D_k|$.

\section{Empirical Evaluation}
\vspace{-1mm}%
\subsection{Data Description}
\vspace{-1mm}%
In this paper, we use the MIMIC-III~(version 1.4) clinical dataset provided by PhysioNet~\cite{PhysioNet}. This dataset provides de-identified data collected from ICU patients in Beth Israel Deaconess Medical Center in Boston, Massachusetts. The data was collected from $46,476$~patients during the years 2001 to 2012 \cite{MIMIC-III,MIMIC-IIIv1.4}. The database includes patient information such as demographics, vital sign measurements, and laboratory test results.

Initially, we select the patients based on the criteria described in \secref{sec:data-preparation}, and in addition, we dismiss patients from the Neonatal Intensive Care Unit (NICU) and Pediatric Intensive Care Unit (PICU).
For this study, we use two different cohorts of patients:
\textbf{(i)~Cohort 1} is identified by $\Delta t_{min}=24\,\mathrm{h}$ and $\Delta t_{max}=72\,\mathrm{h}$, and we compare different $\Delta t_{data} \in [8\,\mathrm{h}, 16\,\mathrm{h}, 24\,\mathrm{h}]$;
and \textbf{(ii)~Cohort 2} is identified by $\Delta t_{min}=48\,\mathrm{h}$ and $\Delta t_{max}=96\,\mathrm{h}$, and we compare $\Delta t_{data} \in [8\,\mathrm{h}, 16\,\mathrm{h}, 24\,\mathrm{h}, 32\,\mathrm{h}, 40\,\mathrm{h}, 48\,\mathrm{h}$].
Both of these cohorts use $\Delta t_{max}=\Delta t_{min}+48\,\mathrm{h}$. For simplicity, we therefore, refer to the cohorts by their $\Delta t_{min}$ only.

For the pre-processing and feature selection, we follow the approach from \cite{Mondrejevski2022,Pattalung2021}. Initially, we extract demographic information, such as gender and age, that is used for describing the cohorts. We also extract $7$ vital signs and $16$ lab values, shown in \tabref{tab:vitals+labs}, in the form of MTS.  In this paper, vital signs are re-sampled in $1\,\mathrm{h}$ intervals, while we use a sampling interval of $8\,\mathrm{h}$ for lab values.%

\vspace{-1mm}
\begin {table} [h!]
\vspace{-6mm}%
\caption{ Vital and Laboratory Values}\label{tab:vitals+labs}%
\centering%
\renewcommand{\arraystretch}{1.2}%
\begin{tabular}{rl}
\toprule[1.5pt]

\minipage{2.5cm}
    \head{Vital Signs:}
    
    {\scriptsize (\textit{1 per hour})}
\endminipage & 
\minipage{8cm}
    \noindent\scriptsize\justifying
    heart-rate, systolic~blood-pressure, diastolic~blood-pressure, mean~blood-pressure, respiratory~rate, core~temperature, blood~oxigen~saturation~(\texttt{spo2})
\endminipage \\

\midrule

\minipage{2.5cm}
    \head{Lab Values:}
    
    {\scriptsize (\textit{1 per 8 hours})}
\endminipage & 
\minipage{8cm}
    \noindent\scriptsize\justifying
    albumin, blood~urea~nitrogen~(\texttt{bun}), bilirubin, lactate, bicarbonate, band~neutrophils~(\texttt{bands}), chloride, creatinine, glucose, hemoglobin, hematocrit, platelets, potassium, partial~thromboplastin~time~(\texttt{ptt}), sodium, white~blood-cells
\endminipage \\

\bottomrule[1.5pt] 
\end{tabular}
\vspace{-2mm}
\end{table}%
Finally, we extract the label indicating whether a patient died or not based on the column \texttt{deathtime} in table \texttt{admissions} of MIMIC-III. If a \texttt{deathtime} is recorded for patient $i$ during the ICU stay, we assign the label $y_i=1$. Otherwise, we assume that the patient survived the ICU and assign the label $y_i=0$. 
As shown in \tabref{tab:cohorts}, there is a heavily imbalanced class distribution as there are more patients in the \textit{discharge} class than in the \textit{death} class.
\vspace{-1mm}
\begin{table} [h!]
\vspace{-6mm}%
\caption{Cohort Sizes and Class Distribution}\label{tab:cohorts}%
\centering%
\setlength{\tabcolsep}{5pt}%
\resizebox{\textwidth}{!}{
\begin{tabular}{l||rr|rr|r||rr|rr|r}
\toprule[1.5pt]
\multirow{3}{*}{} & 
\multicolumn{5}{c||}{\head{Cohort 1 [$\Delta t_{min} = 24\,\mathrm{h}$]}} & 
\multicolumn{5}{c}{\head{Cohort 2 [$\Delta t_{min} = 48\,\mathrm{h}$]}}
\\

&
\multicolumn{2}{c|}{\head{Deaths}} & 
\multicolumn{2}{c|}{\head{Discharges}} & 
\multirow{2}{*}{\head{Total}} &
\multicolumn{2}{c|}{\head{Deaths}} & 
\multicolumn{2}{c|}{\head{Discharges}} & 
\multirow{2}{*}{\head{Total}}\\
&
\scriptsize{absolute} &
\scriptsize{percent} &
\scriptsize{absolute} &
\scriptsize{percent} &
&
\scriptsize{absolute} &
\scriptsize{percent} &
\scriptsize{absolute} &
\scriptsize{percent} &
\\

\midrule[1.5pt]
\head{Patients}         & $   804$ & $  4.4\%$ & $17,477$ & $ 95.6\%$ & $18,281$ 
                        & $   547$ & $  5.4\%$ & $ 9,496$ & $ 94.6\%$ & $10,043$ \\

\midrule
\head{Male}             & $   420$ & $  2.3\%$ & $10,075$ & $ 55.1\%$ & $10,495$ 
                        & $   287$ & $  2.9\%$ & $ 5,255$ & $ 52.3\%$ & $ 5,542$ \\
\head{Female}           & $   384$ & $  2.1\%$ & $ 7,402$ & $ 40.5\%$ & $ 7,786$ 
                        & $   260$ & $  2.6\%$ & $ 4,241$ & $ 42.2\%$ & $ 4,501$ \\

\midrule
\head{Age  0 to 29}     & $    18$ & $  0.1\%$ & $   892$ & $  4.9\%$ & $   910$ 
                        & $    19$ & $  0.2\%$ & $   421$ & $  4.2\%$ & $   440$ \\
\head{Age 30 to 59}     & $   177$ & $  1.0\%$ & $ 5,777$ & $ 31.6\%$ & $ 5,954$ 
                        & $   133$ & $  1.3\%$ & $ 2,863$ & $ 28.5\%$ & $ 2,996$ \\
\head{Age 60 to 89}     & $   525$ & $  2.9\%$ & $ 9,920$ & $ 54.3\%$ & $10,445$ 
                        & $   352$ & $  3.5\%$ & $ 5,692$ & $ 56.7\%$ & $ 6,044$ \\
\head{Age 90+}          & $    84$ & $  0.5\%$ & $   888$ & $  4.9\%$ & $   972$ 
                        & $    43$ & $  0.4\%$ & $   520$ & $  5.2\%$ & $   563$ \\

\bottomrule[1.5pt] 
\end{tabular}
}
\vspace{-6mm}
\end{table}%
%
\subsection{Model Training and Evaluation}%
\vspace{-1mm}%
In this study, we use $F=5$ folds for stratified cross-validation, where we evaluate CML, LML, and FL on the same testing data splits within each cross-validation round. In each round, the data of the remaining four folds are again partitioned in LML and FL models, as those are trained with different numbers of clients~$K \in [2,4,8]$, whilst all remaining data are used in CML. The available data in each scenario (either all remaining data in CML or each client's data in LML and FL) are split into $80\%$~training and $20\%$~validation (see \figref[c]{fig:method}).


We use Adaptive Moment Estimation~(ADAM) \cite{Kingma2014} as an optimizer for updating the network weights in CML, FL, and LML. Additionally, we apply an initial learning rate $\eta$ of $0.01$, which is reduced by $50\%$ every five epochs.
Lastly, we use ES via monitoring the loss or F1-score on the validation set with patience~$P=30$ and the maximum number of epochs set to $100$. In case the F1-score is undefined, i.e., recall and precision are zero, we set it to $-1$. 
The local minibatch size $B$ depends on the number of clients $K$ participating in each configuration: we use $B=512/K$ for performance reasons.
For FL, we use $E=1$ number of local epochs. The fraction of clients computing in each FL round is $C=1$, meaning that all clients participate in each iteration.

To compare the performance of the different settings, we use the evaluation metrics AUPRC, AUROC, Precision, Recall, and F1-score. AUROC is chosen as it is commonly used to assess the performance of ICU mortality prediction.
However, AUPRC and F1-score are more suitable for highly imbalanced classes, which is the case for our problem. 

The entire code produced in this paper is publicly available on GitHub \footnote{\url{https://github.com/randlbem/Early_ICU_mortality_prediction_with_deep_FL.git}}. 

\vspace{-2mm}%
\subsection{Results \& Discussion}%
\vspace{-1mm}%
As previously described, we evaluate the ability of our proposed FL workflow to predict the risk of ICU mortality at an early stage. We compare it with the CML and LML approaches on two cohorts of patients ($\Delta t_{min} = 24\,\mathrm{h}$ and $48\,\mathrm{h}$), using two ES metrics (loss and F1-score) and different time windows $\Delta t_{data}$.

The results for cohort $\Delta t_{min}=24\,\mathrm{h}$, using the minimum loss for ES, are shown in \tabref{tab:PerformanceCohort1}. Overall, the results show that our model performs well in the task of early ICU mortality prediction. For example, for FL with $\Delta t_{data}=24\,\mathrm{h}$, we obtain an average AUROC of $0.90 \pm 0.01$ and an AUPRC of  $0.47 \pm 0.04$. It's important to highlight that due to the class imbalance of $4.4\,\%$ towards the positive class, the baseline value for the AUPRC is $0.044$.

In addition, we yield three conclusions from \tabref{tab:PerformanceCohort1}: 
\textbf{(i)}~While an increasing number of clients results in a decrease in predictive performance over all the metrics in LML, the performance of FL remains close to that of CML, regardless of the number of clients.
\textbf{(ii)}~With growing $\Delta t_{data}$, and respectively shrinking $\Delta t_{pred}$, the model performance increases. 
\textbf{(iii)}~The relation between precision and recall is very unstable, which is shown in the fluctuation of the averages and the high standard deviations.
While (i) clearly shows that FL has the potential to improve on LML for early ICU mortality prediction, (ii) and (iii) are further explored in the following experiments.%
\begin{table} [h]%
\vspace{-3mm}%
\centering%
\caption{ES with min. loss ($\Delta t_{min}=24\,\mathrm{h}$).}\label{tab:PerformanceCohort1}%
\setlength{\tabcolsep}{5pt}%
\resizebox{\textwidth}{!}{
\begin{tabular}{r||r|rrr|rrr}
\toprule[1.5pt]

\multicolumn{1}{c||}{\head{score}}
&\multicolumn{1}{c|}{\head{CML}}
&\multicolumn{3}{c|}{\head{FL}}
&\multicolumn{3}{c}{\head{LML}}
\\
&
&\multicolumn{1}{c}{\scriptsize{$2~clients$}}
&\multicolumn{1}{c}{\scriptsize{$4~clients$}}
&\multicolumn{1}{c|}{\scriptsize{$8~clients$}}
&\multicolumn{1}{c}{\scriptsize{$2~clients$}}
&\multicolumn{1}{c}{\scriptsize{$4~clients$}}
&\multicolumn{1}{c}{\scriptsize{$8~clients$}}
\\

\midrule[1.5pt]
\multicolumn{8}{l}{$\bf\Delta t_{data} = 8\,\mathrm{h}$; avg. $\Delta t_{pred} = 35.2\,\mathrm{h}$} \\
\midrule
\head{AUROC}	    & $0.87 \pm 0.02$	& $0.86 \pm 0.01$	& $0.86 \pm 0.01$	& $0.87 \pm 0.01$	& $0.85 \pm 0.01$	& $0.82 \pm 0.01$	& $0.81 \pm 0.01$	\\
\head{AUPRC}	    & $0.36 \pm 0.03$	& $0.37 \pm 0.02$	& $0.37 \pm 0.02$	& $0.37 \pm 0.03$	& $0.34 \pm 0.04$	& $0.31 \pm 0.03$	& $0.29 \pm 0.02$	\\
\head{F1}	        & $0.29 \pm 0.15$	& $0.38 \pm 0.01$	& $0.36 \pm 0.05$	& $0.38 \pm 0.04$	& $0.34 \pm 0.07$	& $0.34 \pm 0.02$	& $0.25 \pm 0.04$	\\
\head{precision}	& $0.62 \pm 0.21$	& $0.38 \pm 0.07$	& $0.39 \pm 0.11$	& $0.41 \pm 0.08$	& $0.53 \pm 0.14$	& $0.41 \pm 0.06$	& $0.48 \pm 0.08$	\\
\head{recall}	    & $0.24 \pm 0.14$	& $0.41 \pm 0.07$	& $0.42 \pm 0.14$	& $0.39 \pm 0.09$	& $0.29 \pm 0.10$	& $0.30 \pm 0.07$	& $0.17 \pm 0.04$	\\

\midrule[1.5pt]
\multicolumn{8}{l}{$\bf\Delta t_{data} = 16\,\mathrm{h}$; avg. $\Delta t_{pred} = 27.2\,\mathrm{h}$} \\
\midrule
\head{AUROC}	    & $0.89 \pm 0.01$	& $0.88 \pm 0.01$	& $0.88 \pm 0.01$	& $0.88 \pm 0.01$	& $0.87 \pm 0.01$	& $0.85 \pm 0.01$	& $0.82 \pm 0.01$	\\
\head{AUPRC}	    & $0.44 \pm 0.04$	& $0.41 \pm 0.04$	& $0.41 \pm 0.04$	& $0.42 \pm 0.03$	& $0.40 \pm 0.04$	& $0.35 \pm 0.05$	& $0.32 \pm 0.02$	\\
\head{F1}	        & $0.42 \pm 0.01$	& $0.35 \pm 0.07$	& $0.41 \pm 0.02$	& $0.38 \pm 0.08$	& $0.29 \pm 0.05$	& $0.29 \pm 0.12$	& $0.22 \pm 0.02$	\\
\head{precision}	& $0.51 \pm 0.10$	& $0.58 \pm 0.13$	& $0.47 \pm 0.05$	& $0.46 \pm 0.07$	& $0.55 \pm 0.18$	& $0.50 \pm 0.14$	& $0.55 \pm 0.09$	\\
\head{recall}	    & $0.38 \pm 0.07$	& $0.27 \pm 0.09$	& $0.37 \pm 0.04$	& $0.34 \pm 0.10$	& $0.21 \pm 0.03$	& $0.22 \pm 0.10$	& $0.14 \pm 0.02$	\\

\midrule[1.5pt]
\multicolumn{8}{l}{$\bf\Delta t_{data} = 24\,\mathrm{h}$; avg. $\Delta t_{pred} = 19.2\,\mathrm{h}$} \\
\midrule
\head{AUROC}	    & $0.89 \pm 0.01$	& $0.90 \pm 0.01$	& $0.90 \pm 0.01$	& $0.89 \pm 0.01$	& $0.89 \pm 0.01$	& $0.87 \pm 0.01$	& $0.83 \pm 0.01$	\\
\head{AUPRC}	    & $0.48 \pm 0.03$	& $0.48 \pm 0.03$	& $0.47 \pm 0.04$	& $0.45 \pm 0.04$	& $0.46 \pm 0.03$	& $0.41 \pm 0.05$	& $0.37 \pm 0.03$	\\
\head{F1}	        & $0.10 \pm 0.11$	& $0.33 \pm 0.17$	& $0.34 \pm 0.07$	& $0.30 \pm 0.17$	& $0.22 \pm 0.12$	& $0.26 \pm 0.07$	& $0.32 \pm 0.05$	\\
\head{precision}	& $0.66 \pm 0.38$	& $0.70 \pm 0.18$	& $0.75 \pm 0.06$	& $0.72 \pm 0.16$	& $0.51 \pm 0.18$	& $0.47 \pm 0.10$	& $0.59 \pm 0.15$	\\
\head{recall}	    & $0.05 \pm 0.07$	& $0.27 \pm 0.16$	& $0.22 \pm 0.05$	& $0.23 \pm 0.14$	& $0.15 \pm 0.08$	& $0.19 \pm 0.06$	& $0.23 \pm 0.05$	\\

\bottomrule[1.5pt] 
\end{tabular}
}
\vspace{-2mm}
\end{table}%
\vspace{-2mm}%
\subsubsection{Studying the influence of the size of $\Delta t_{data}$ and $\Delta t_{pred}$.}
\begin{figure}[htbp!]%
\begin{tikzpicture}%
\draw (0,.5) node[anchor=north]{\includegraphics[trim = .4cm .2cm .4cm .2cm, clip, width=.9\textwidth]{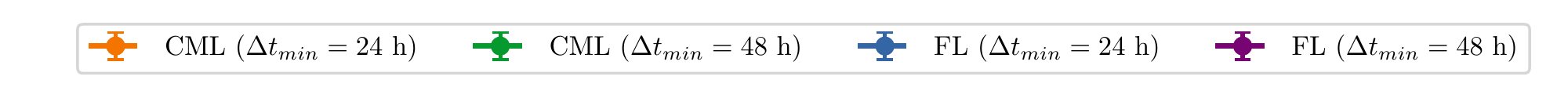}};%
\draw (-3,0) node[anchor=north]{\includegraphics[width=5.8cm]{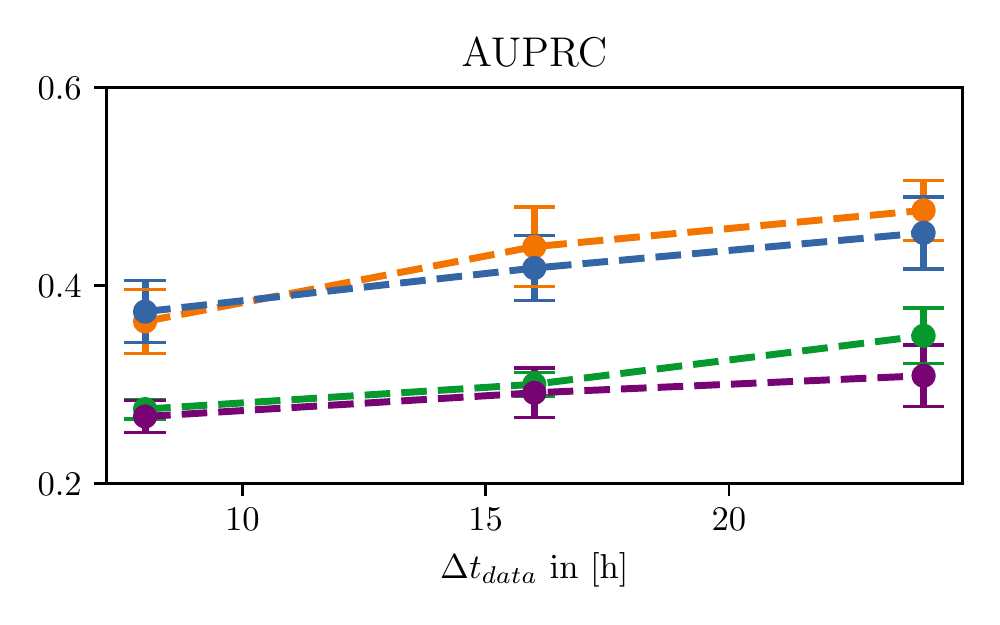}};%
\draw (3,0) node[anchor=north]{\includegraphics[width=5.8cm]{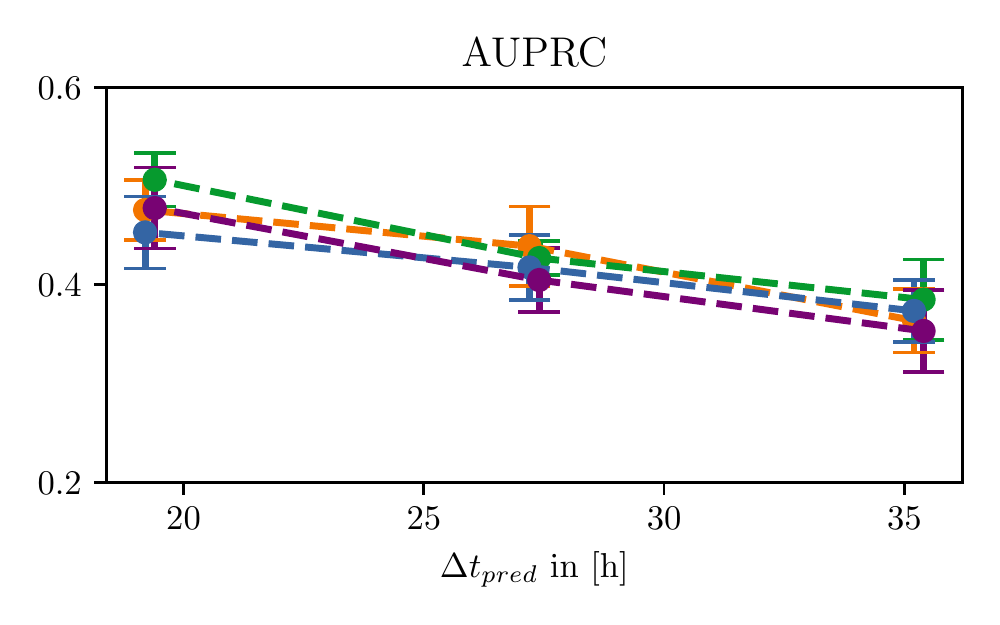}};%
\draw (-5,-.6) node[anchor=north]{\footnotesize{\color{gray}(a)}};%
\draw (5.4,-.6) node[anchor=north]{\footnotesize{\color{gray}(b)}};%
\end{tikzpicture}%

\noindent\justifying\scriptsize{The dots represent the mean values over the 5-fold cross-validation iterations. The vertical bars show the standard deviation. The scores were calculated on the test sets, while the ES metric is the loss.}%
\caption{Mean AUPRC of the two scenarios ($\Delta t_{min}=24\,\mathrm{h}$ and $\Delta t_{min}=48\,\mathrm{h}$).}%
\label{fig:results1}%
\vspace{-2mm}
\end{figure}

\vspace{-2mm}%
To assess whether the performance improves with an increasing $\Delta t_{data}$, decreases with an increasing $\Delta t_{pred}$, or both, we compare the test scores of the two cohorts ($\Delta t_{min}=24\,\mathrm{h}$ and $\Delta t_{min}=48\,\mathrm{h}$). The comparison is shown in \figref{fig:results1}. 
The figure shows that cohort $\Delta t_{min}=24\,\mathrm{h}$ achieves higher performance than cohort $\Delta t_{min}=48\,\mathrm{h}$, at the same $\Delta t_{data}$ (\figref[a]{fig:results1}). Nevertheless, both cohorts' performance increases alongside $\Delta t_{data}$. When comparing the performance over the length of $\Delta t_{pred}$, we see that with a rising $\Delta t_{pred}$, all the curves are decreasing in a similar manner (\figref[b]{fig:results1}). 
This behavior can be seen over different models and metrics and is a strong indicator that the size of $\Delta t_{pred}$ is more important than the size of $\Delta t_{data}$, since $\Delta t_{pred}$ is bigger in cohort $\Delta t_{min}=48\,\mathrm{h}$ than cohort  $\Delta t_{min}=24\,\mathrm{h}$  for the same $\Delta t_{data}$. This means that even for small $\Delta t_{data}$, $8\,\mathrm{h}$ or just $6\,\mathrm{h}$ as Awad et al.~\cite{Awad2020} demonstrate, prediction should be possible, if $\Delta t_{pred}$ is small enough.

\begin{table} [h!]%
\vspace{-3mm}%
\centering%
\caption{ES with max. F1 ($\Delta t_{min}=24\,\mathrm{h}$).}\label{tab:PerformanceCohort1_ES-F1}%
\setlength{\tabcolsep}{5pt}%
\resizebox{\textwidth}{!}{
\begin{tabular}{r||r|rrr|rrr}
\toprule[1.5pt]

\multicolumn{1}{c||}{\head{score}}
&\multicolumn{1}{c|}{\head{CML}}
&\multicolumn{3}{c|}{\head{FL}}
&\multicolumn{3}{c}{\head{LML}}
\\
&
&\multicolumn{1}{c}{\scriptsize{$2~clients$}}
&\multicolumn{1}{c}{\scriptsize{$4~clients$}}
&\multicolumn{1}{c|}{\scriptsize{$8~clients$}}
&\multicolumn{1}{c}{\scriptsize{$2~clients$}}
&\multicolumn{1}{c}{\scriptsize{$4~clients$}}
&\multicolumn{1}{c}{\scriptsize{$8~clients$}}
\\

\midrule[1.5pt]
\multicolumn{8}{l}{$\bf\Delta t_{data} = 8\,\mathrm{h}$; avg. $\Delta t_{pred} = 35.2\,\mathrm{h}$} \\
\midrule
\head{AUROC}	    & $0.87 \pm 0.01$	& $0.87 \pm 0.01$	& $0.87 \pm 0.01$	& $0.87 \pm 0.01$	& $0.86 \pm 0.01$	& $0.84 \pm 0.01$	& $0.80 \pm 0.01$	\\
\head{AUPRC}	    & $0.39 \pm 0.04$	& $0.38 \pm 0.04$	& $0.37 \pm 0.04$	& $0.37 \pm 0.03$	& $0.36 \pm 0.05$	& $0.32 \pm 0.04$	& $0.28 \pm 0.02$	\\
\head{F1}	        & $0.40 \pm 0.03$	& $0.38 \pm 0.02$	& $0.39 \pm 0.04$	& $0.39 \pm 0.04$	& $0.39 \pm 0.03$	& $0.36 \pm 0.03$	& $0.35 \pm 0.02$	\\
\head{precision}	& $0.42 \pm 0.07$	& $0.38 \pm 0.05$	& $0.35 \pm 0.06$	& $0.35 \pm 0.08$	& $0.37 \pm 0.05$	& $0.37 \pm 0.05$	& $0.35 \pm 0.04$	\\
\head{recall}	    & $0.41 \pm 0.08$	& $0.41 \pm 0.11$	& $0.46 \pm 0.08$	& $0.47 \pm 0.09$	& $0.43 \pm 0.04$	& $0.36 \pm 0.03$	& $0.34 \pm 0.03$	\\

\midrule[1.5pt]
\multicolumn{8}{l}{$\bf\Delta t_{data} = 16\,\mathrm{h}$; avg. $\Delta t_{pred} = 27.2\,\mathrm{h}$} \\
\midrule
\head{AUROC}	    & $0.89 \pm 0.01$	& $0.88 \pm 0.01$	& $0.88 \pm 0.01$	& $0.88 \pm 0.01$	& $0.87 \pm 0.01$	& $0.85 \pm 0.01$	& $0.82 \pm 0.01$	\\
\head{AUPRC}	    & $0.44 \pm 0.05$	& $0.41 \pm 0.04$	& $0.41 \pm 0.04$	& $0.42 \pm 0.03$	& $0.40 \pm 0.05$	& $0.37 \pm 0.04$	& $0.33 \pm 0.03$	\\
\head{F1}	        & $0.44 \pm 0.02$	& $0.41 \pm 0.01$	& $0.40 \pm 0.05$	& $0.43 \pm 0.03$	& $0.41 \pm 0.02$	& $0.39 \pm 0.03$	& $0.37 \pm 0.02$	\\
\head{precision}	& $0.46 \pm 0.05$	& $0.41 \pm 0.04$	& $0.43 \pm 0.06$	& $0.43 \pm 0.04$	& $0.38 \pm 0.08$	& $0.38 \pm 0.05$	& $0.41 \pm 0.03$	\\
\head{recall}	    & $0.42 \pm 0.05$	& $0.42 \pm 0.07$	& $0.40 \pm 0.11$	& $0.43 \pm 0.06$	& $0.46 \pm 0.08$	& $0.42 \pm 0.06$	& $0.33 \pm 0.04$	\\

\midrule[1.5pt]
\multicolumn{8}{l}{$\bf\Delta t_{data} = 24\,\mathrm{h}$; avg. $\Delta t_{pred} = 19.2\,\mathrm{h}$} \\
\midrule
\head{AUROC}	    & $0.90 \pm 0.01$	& $0.90 \pm 0.01$	& $0.90 \pm 0.00$	& $0.89 \pm 0.01$	& $0.89 \pm 0.01$	& $0.87 \pm 0.01$	& $0.83 \pm 0.01$	\\
\head{AUPRC}	    & $0.50 \pm 0.04$	& $0.49 \pm 0.04$	& $0.47 \pm 0.04$	& $0.47 \pm 0.03$	& $0.47 \pm 0.04$	& $0.43 \pm 0.04$	& $0.37 \pm 0.03$	\\
\head{F1}	        & $0.49 \pm 0.04$	& $0.48 \pm 0.03$	& $0.46 \pm 0.03$	& $0.44 \pm 0.03$	& $0.45 \pm 0.02$	& $0.42 \pm 0.02$	& $0.40 \pm 0.02$	\\
\head{precision}	& $0.55 \pm 0.05$	& $0.52 \pm 0.05$	& $0.49 \pm 0.07$	& $0.45 \pm 0.07$	& $0.48 \pm 0.08$	& $0.39 \pm 0.04$	& $0.46 \pm 0.02$	\\
\head{recall}	    & $0.46 \pm 0.08$	& $0.46 \pm 0.05$	& $0.44 \pm 0.05$	& $0.46 \pm 0.11$	& $0.46 \pm 0.10$	& $0.48 \pm 0.09$	& $0.35 \pm 0.03$	\\

\bottomrule[1.5pt] 
\end{tabular}
}
\end{table}%

\vspace{-3mm}%
\subsubsection{Stabilizing precision \& recall.}%
\begin{figure}[t]%
\noindent\centering\includegraphics[trim = 0.1cm .2cm 0.1cm .4cm, clip, width=\textwidth]{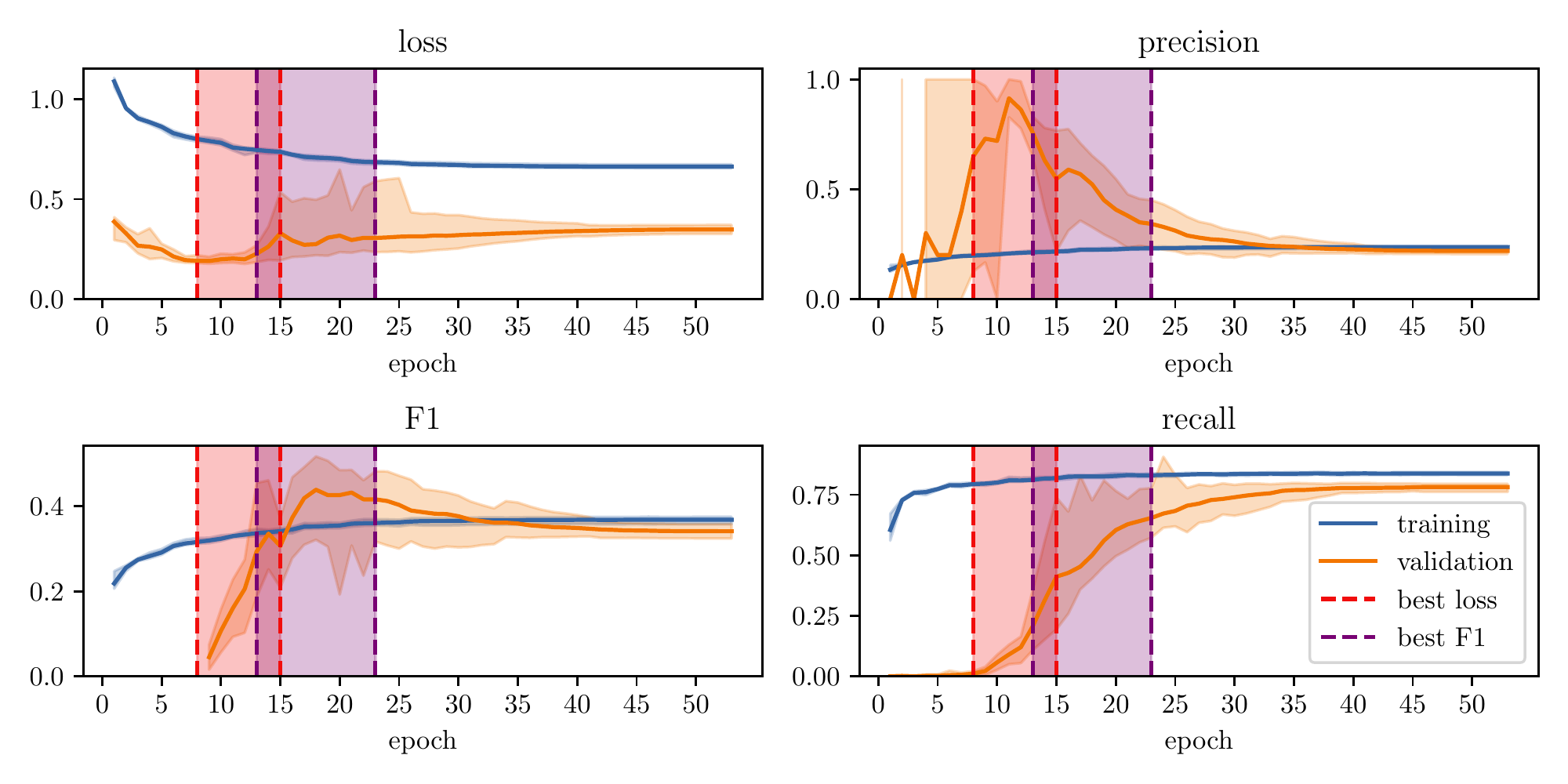}%

\noindent\justifying\scriptsize{The curves represent the mean values of the validation scores over the 5-fold cross-validation iterations using loss as the ES metric. The borders of the shaded areas mark the standard deviation. The best loss and F1 values recognized by ES fall within the red and purple regions.}%
\caption{Learning progress of CML ($\Delta t_{min}=24\,\mathrm{h}$).}%
\label{fig:results2}%
\vspace{-2mm}
\end{figure}%

To better understand the interplay between precision and recall, we examine their fluctuations from the CML model, as it represents the most basic case. Its learning curve in \figref{fig:results2} shows that there is a trade-off between precision and recall: precision increases during the first 12 epochs and then shows a decline, while recall increases steadily. We can also see that the loss shows a very flat minimum while precision is still stabilizing, and recall has just begun to increase. This means that minimal changes in the loss's progression can greatly impact the models' precision and recall. The F1-score, however, shows a more defined maximum.

In order to stabilize precision and recall, we re-train the models whose performance is shown in \tabref{tab:PerformanceCohort1}, using the F1-score as the ES metric. As the F1-Score is the harmonic mean between precision and recall, using it for stopping the training should create a model with an optimal balance between precision and recall. The results are shown in \tabref{tab:PerformanceCohort1_ES-F1}. A comparison of tables \tabref{tab:PerformanceCohort1} and \tabref{tab:PerformanceCohort1_ES-F1} shows that ES with the highest F1-score produces better and more stable results for all models. For example, for FL with $\Delta t_{data}=24\,\mathrm{h}$, we now obtain an average F1-score of $0.46 \pm 0.03$, instead of $0.32 \pm 0.15$ (as shown in \tabref{tab:PerformanceCohort1}) in addition to marginally higher AUROC and AUPRC values.

\vspace{-2mm}%
\section{Conclusion}%
\vspace{-2mm}%
We present an FL workflow that allows for \textit{early ICU mortality prediction}. Our results show that FL performs equally well as the CML approach and substantially better than the LML, especially as the number of clients increases. While the performance remains stable in FL with 2, 4, and 8 clients, in LML, the performance decreases considerably with an increasing number of clients. These findings are based on the AUROC score - widely used in the literature but ill-suited for the heavily imbalanced data in this problem - but also on the more meaningful AUPRC and F1-score.

Furthermore, our results indicate, in agreement with literature \cite{Awad2020,Pattalung2021}, that the size of the prediction window is much more important for the performance in early prediction tasks than the length of patient history during which data is collected. Thus, the results show better predictive performance when the patient history window is closer to the end of the ICU stay. Lastly, we show that using the F1-score as an ES metric can stabilize and increase the predictive performance in tasks like ours.

Nevertheless, this study also creates the basis for future work. Since we limit ourselves to horizontal and stratified client splits for comparability reasons, it is necessary to re-evaluate our findings in more realistic settings. In addition, our FL workflow performs considerably well in predicting ICU mortality at an early stage using the MIMIC-III dataset. However, the generalizability of the approach needs to be tested beyond this specific dataset.
Furthermore, it would be interesting to explore how far ahead of death or discharge \textit{early ICU mortality} can be reasonably predicted and hereby expand on our findings.

%
\vspace{-2mm}%
\bibliographystyle{splncs04}%
\bibliography{sources}%

\begin{thebibliography}{10}
\providecommand{\url}[1]{\texttt{#1}}
\providecommand{\urlprefix}{URL }
\providecommand{\doi}[1]{https://doi.org/#1}

\bibitem{covid2022}
Auld, S.C., Harrington, K.R.V., Adelman, M.W., Robichaux, C.J., Overton, E.C.,
  Caridi-Scheible, M., Coopersmith, C.M., Murphy, D.J.: Research collaborative
  trends in icu mortality from coronavirus disease 2019: A tale of three
  surges. the Emory COVID-19 Quality and Clinical  \textbf{50}(2),  245--255
  (2022)

\bibitem{Awad2020}
Awad, A., Bader-El-Den, M., Briggs, J., McNicholas, J., El-Sonbaty, Y.:
  Predicting hospital mortality for intensive care unit patients: Time-series
  analysis. Health Informatics Journal  \textbf{26}(2),  1043--1059 (2020)

\bibitem{Capuzzo2014}
Capuzzo, M., Volta, C., Tassinati, T., Moreno, R., Valentin, A., Guidet, B.,
  Iapichino, G., Martin, C., Perneger, T., Combescure, C., Poncet, A., Rhodes,
  A., Oeyen, S., Matejovic, M., Toft, P., Wrigge, H., et~al.: Hospital
  mortality of adults admitted to intensive care units in hospitals with and
  without intermediate care units: A multicentre european cohort study.
  Critical Care  \textbf{18}(5) (2014)

\bibitem{Cho2014}
Cho, K., van Merrienboer, B., Bahdanau, D., Bengio, Y.: On the properties of
  neural machine translation: Encoder-decoder approaches (2014)

\bibitem{Gall1993}
Gall, J.R., Lemeshow, S., Saulnier, F.: A new simplified acute physiology score
  (saps ii) based on a european/north american multicenter study. JAMA: The
  Journal of the American Medical Association  \textbf{270}(24),  2957--2963
  (1993)

\bibitem{Wendong2018}
Ge, W., Huh, J.W., Rang, Y.P., Lee, J., Kim, Y.H., Turchin, A.: An
  interpretable icu mortality prediction model based on logistic regression and
  recurrent neural networks with lstm units. In: AMIA annual symposium. vol.~1,
  pp. 460 -- 469. American Medical Informatics Association (2018)

\bibitem{PhysioNet}
Goldberger, A.L., Amaral, L.A., Glass, L., Hausdorff, J.M., Ivanov, P.C., Mark,
  R.G., Mietus, J.E., Moody, G.B., Peng, C.K., Stanley, H.E.: Physiobank,
  physiotoolkit, and physionet: components of a new research resource for
  complex physiologic signals. Circulation  \textbf{101}(23),  E215--220 (2000)

\bibitem{Hochreiter1997}
Hochreiter, S., Schmidhuber, J.: Long short-term memory. Neural Computation
  \textbf{9}(8), ~1735 (1997)

\bibitem{MIMIC-III}
Johnson, A.E.W., Pollard, T.J., Lehman, L.W.H., Feng, M., Ghassemi, M., Moody,
  B., Celi, L.A., Mark, R.G., Shen, L., Szolovits, P.: Mimic-iii, a freely
  accessible critical care database. Scientific Data  \textbf{3} (2016)

\bibitem{MIMIC-IIIv1.4}
Johnson, A.E.W., Pollard, T.J., Mark, R.G.: Mimic-iii clinical database
  (version 1.4). PhysioNet  (2016)

\bibitem{Johnson2017}
Johnson, A.E., Mark, R.G.: Real-time mortality prediction in the intensive care
  unit. In: AMIA Annual Symposium Proceedings. vol.~2017, p.~994. American
  Medical Informatics Association (2017)

\bibitem{Kingma2014}
Kingma, D.P., Ba, J.: Adam: A method for stochastic optimization. arXiv
  preprint arXiv:1412.6980  (2014)

\bibitem{Knaus1991}
Knaus, W., Wagner, D., Draper, E., Zimmerman, J., Bergner, M., Bastos, P.,
  Sirio, C., Murphy, D., Lotring, T., Damiano, A., Harrell~Jr., F.: The apache
  iii prognostic system: Risk prediction of hospital mortality for critically
  iii hospitalized adults. Chest  \textbf{100}(6),  1619--1636 (1991)

\bibitem{McMahan2016}
McMahan, H.B., Moore, E., Ramage, D., Hampson, S., Arcas, B.A.y.:
  Communication-efficient learning of deep networks from decentralized data
  (2016)

\bibitem{Mondrejevski2022}
Mondrejevski, L., Miliou, I., Montanino, A., Pitts, D., Hollmen, J.,
  Papapetrou, P.: Flicu: A federated learning workflow for intensive care unit
  mortality prediction. In: 2022 IEEE 35th International Symposium on
  Computer-Based Medical Systems (CBMS). pp. 32--37. IEEE Computer Society, Los
  Alamitos, CA, USA (2022)

\bibitem{Pattalung2021}
Na~Pattalung, T., Ingviya, T., Chaichulee, S.: Feature explanations in
  recurrent neural networks for predicting risk of mortality in intensive care
  patients. Journal of Personalized Medicine  \textbf{11}(9) (2021)

\bibitem{Shokri2015}
Shokri, R., Shmatikov, V.: Privacy-preserving deep learning. In: Proceedings of
  the 22nd ACM SIGSAC Conference on Computer and Communications Security. p.
  1310–1321. CCS '15, Association for Computing Machinery, New York, NY, USA
  (2015)

\end{thebibliography}
\end{document}